\documentclass[11pt]{article}

\usepackage{amsmath, amsfonts, amsthm}
\usepackage{graphics, epsfig, epstopdf}
\usepackage{multirow}
\usepackage[authoryear]{natbib}

\def\ba{\mathbf a}

\title{A new approach for physiological time series}
\author{Dong Mao\\
AMSL Brion, 4211 Burton Drive, \\
Santa Clara, CA 95054, USA
\\
\\
Yang Wang\\
Department of Mathematics, \\
Hong Kong University of Science and Technology, \\
Hong Kong\\
\\
Qiang Wu\\
Department of Mathematical Sciences,\\
Middle Tennessee State University, \\
1301 E Main Street,
Murfreesboro, TN 37132, USA}

\begin{document}
\maketitle

\begin{abstract}

We developed a new approach for the analysis of physiological time series. 
An iterative convolution filter is used to decompose the time series into various components. Statistics of 
these components  are extracted as features
to characterize the mechanisms underlying the time series.
Motivated by the studies that  show many normal physiological systems  involve 
irregularity  while the decrease of irregularity usually implies the abnormality,
the statistics for ``outliers" in the components are used as features measuring irregularity.
Support vector machines are used to select the most relevant features 
that are able to differentiate the time series from normal and abnormal systems.
This new approach is successfully used in the study of congestive heart failure 
by heart beat interval time series.

\bigskip

\end{abstract}

\newpage

\section{Introduction}

An understanding of physiological time series such as the heart-beat
intervals is important to many areas, like heart-attack prediction,
cardiovascular health, sport and exercise, etc. The study of time
series can reveal underlying mechanisms of the physiological system,
which usually contains both deterministic and stochastic components.
Therefore the analysis of time series is very complicated because of
the nonlinear and non-stationary characteristics of physiological
time series data. Over the past years, time series analysis methods
are applied to quantify physiological data for identification and
classification \citep{Kantz, Schreiber}. The applications of
physiological time series analysis commonly focus on measuring
different aspects of time series data such as complexity,
regularity, predictability, dimensionality, randomness, self
similarity, etc. The tools used in these techniques include but not
restrict to the mean, standard deviation, Fourier transform,
wavelet, entropy, fractal dimension, pattern detection \citep{Kantz2,Tong}.

Recently a new mathematical tool, empirical mode decomposition
(EMD), was proposed by Norden Huang and his collaborators \citep{Huang, Huang2}. 
It decomposes a time series into a finite sum of intrinsic
mode functions (IMF) that generally admit well-behaved Hilbert
transforms. This decomposition is based on the local characteristic
time scale of the data, which makes EMD applicable to analyze
nonlinear and non-stationary signals. EMD and Hilbert transform
together, called the Hilbert-Huang transform (HHT), usually allow to
construct meaningful time-frequency representations of signals using
instantaneous frequency of the data. EMD and HHT have been applied
with great success in many application areas such as biological and
medical sciences, geology, astronomy, engineering, and others 
\citep{Huang,Chen,Echeverria,Huang2,Pines,Liu}. Another interesting
set of examples is the work of L.Yang and his collaborators, who have successfully applied
EMD based techniques for texture analysis and Chinese handwriting
recognition \citep{Yang,Yang2,Zheng}.

The main purpose of this paper is to develop a new approach for the
analysis of physiological times series. Our approach is motivated by
two intuitions and coupled with modern machine learning techniques.
The first intuition comes from a belief that a physiological system
should contain a deterministic part that reflects the basic
mechanism for the system to survive and a stochastic part that
represents the variability of resilience. Mathematically they can be
represented by the low frequency and high frequency components of a
physiological signal. This motivates the application of methods of
decomposing signals into various components according to frequencies
in the quantitative analysis of physiological time series. Examples
include the Fourier transform, wavelets, and EMD. In our method we will
use an iterative convolution filter which is an alternative of EMD.
The second intuition comes from a statistical perspective of
irregularity. A lot of study has proved that normal physiological
systems show irregularity due to the existence of stochastic
components while the decrease of irregularity usually implies the
abnormality. From statistical perspective, irregularity of a data
set is represented by the ``outliers". This motivates us to study
the statistics of outliers in physiological time series. However, we
must be careful in doing so. Practical physiological times series
usually contain noise which may also appear as outliers. We have to
guarantee the ``outliers" we examined are not pure noise. This is
possible because true outliers do not have informative structures
and could be detected. The second intuition is the motivation for
our feature construction in Section \ref{sec:feature}.

These two intuitions enable us to decompose the physiological times
series and construct features for our quantitative analysis.
Combining with the well established feature selection techniques in
machine learning we can remove the redundancy of the features and
find relevant statistics for classification of physiological time
series. Support vector machine recursive feature
elimination(SVM-RFE) is suggested in this paper for linear classification
problems. The details of our approach will be described in Section
\ref{sec:method}.

We will use our approach to analyze the heart beat interval time series and study the congestive heart failure problem. 
The study of heart diseases such as congestive heart failure by using heart beat interval times series 
has a long history. Decrease of heart rate variability or cardiac chaos has been 
found in congestive heart failure \citep{poon1997decrease, casolo1989decreased}.
In the literature, many methods and metrics have been proposed to analyze the difference 
between the  heart rate times series of healthy people and congestive heart failure patients,
to name a few, the detrended fluctuation analysis \citep{peng1995quantification}, 
multifractality \citep{ivanov1999multifractality}, multiscaling entropy \citep{med1}, 
hierarchical entropy \citep{jiang2011hierarchical}. 
Our approach is different from the methods in the literature. It incorporates
advanced machine learning techniques and allows the data ``to speak by itself.''
By applying our approach, the purposes are two-fold: The first is to build good
classifiers to enable good diagnosis. The second is to find what kind
of irregularity is associated to the heart health. The results and
discussions are summarized in Section \ref{sec:experiment}.

The novelty of our method is mainly the following two points.
Firstly, although we decompose the time series into components of
different frequencies, we do not compare them from the frequency
domain. Secondly, we proved that the outliers
in a physiological time series are usually not pure noise  but are
informative instead.
 Interestingly, although this idea is motivated 
by physiological times series analysis, it is also found successful
in the stylometry analysis of artworks \citep{hughes2012empirical}.

\section{Method}
\label{sec:method}

\subsection{Signal decomposition}

Let $L$ be a low pass filter. Denote by $T$ the weak limit of the
the operator $(I-L)^n$ as $n\to \infty$, i.e., for a discrete signal
$X$ and time $t$
$$T(X)(t)  = \lim_{n\to \infty} (I - L)^n (X)(t).$$
Using this operator iteratively, a signal $X$ can be decomposed as
follows: Let $F_1 = T(X)$ and for $k\ge 2$,
$$F_k = T\left(X-\sum_{i=1}^{k-1} F_i\right).$$
After $m$ steps we get $F_1, \ldots, F_m$ which we call mode
functions and the residual $$R = X-\sum_{i=1}^m F_i.$$ Then we have
$$ X = F_1 + F_2 + \ldots +F_m +R.$$
In this decomposition, roughly speaking the former mode functions
are noise or high frequency components and the latter mode functions
are low frequency components and $R$ is the trend.

This procedure follows the spirit of the traditional EMD introduced
in \cite{Huang}. In the traditional EMD, the low pass filter $L$ is
chosen as the average of the upper envelope (the cubic spline
connecting the local maxima) and the lower envelope (the cubic
spline connecting the local minima). This method, although has been
successfully used in many applications, is lack of theoretical
foundation and has its limitations.

In \cite{Lin} a new approach is proposed. In this new approach the
low pass filter is a moving average generated by a mask $\ba =
(a_j)_{j=-N}^{N}$ that gives the $L(X)$ as the convolution of $a$
and $X$, i.e.,
$$L(X) (t) = \sum_{j=-N}^N a_j X(j+t).$$
With this choice of $L$ we call the operator $T$ an iterative
convolution filter. A rigorous mathematical foundation and
convergence analysis is given in \cite{Lin, Wang2}. Note the mask $\ba$ is
finitely supported on $[-N, N]$ and $N$ is called the window size.
The flexibility to choose the window size is crucial in applications
and forms a main advantage of this method.

Similar to decompositions by many other methods like Fourier
transform and wavelets, the trend and low frequency components are
usually assumed to characterize the profile of the signal and the
high frequency components characterize the details. In different
applications we need the features of difference components.

\subsection{Feature extraction}
\label{sec:feature}

After decomposing the signal into the mode functions and the trend,
we need to extract statistics that can characterize the essential
features of these components. This step requires a priori knowledge
of the problem under consideration. It could be rather weak. But
without any priori knowledge, it is difficult to get proper
statistics. Also, this step is strongly problem dependent. In the
following let us use the heart-beat intervals as an example to
illustrate how to construct the features.

In this application, each time series is a record of heart beat intervals in 24 hours \citep{med1}. 
It is first decomposed into several mode functions. 
To extract the features, for each mode function $F_i,$ we first get its mean $m_i$ and
standard deviation $\sigma_i$. By the previous studies \citep{poon1997decrease, casolo1989decreased,med1}
the healthy heart beats more irregularly than the unhealthy heart.
This motivates us to design the statistics to measure the
irregularity. To this end, we consider the terms that are larger
than $m_i+\sigma_i$ and find their mean $m_{i,1}$ and standard deviation $\sigma_{i,1}$. We also
find the mean $m_{i,2}$ and standard deviation $\sigma_{i,2}$ of the terms that are larger
than $m_i+2\sigma_i$. Symmetrically we also get the mean $m_{i, -1}$ and standard 
deviation $\sigma_{i,-1}$ of those terms that are smaller than $m_i-\sigma_i$, and
the mean $m_{i, -2}$ and standard 
deviation $\sigma_{i,-2}$ of those terms that are smaller than $m_i-2\sigma_i.$ 
This procedure gives us 10 statistics.
Note all those terms that are outside the one or two standard deviations 
are in some sense ``outliers'' and
it is natural to use their statistics ($m_{i,j}$ and $\sigma_{i,j}$ for $j=1,2,-1,-2$) to characterize the irregularity.
Next we consider the times series $U_i$ composed of local maxima of $F_i$ and 
the time series $L_i$ composed of the local minima of $F_i$. These
two series measure the local amplitude. For each series we
compute the 10 statistics by the same procedure above as for $F_i$. 
Therefore for each mode function $F_i$  we get 30 statistics.

Unlike in \citep{med1}, we use the whole 24-hour heart beat time
series and assume we do not know the periods for different
activities such as sleeping and walking. We think the statistics for
different periods should be different and not all of them represent
the difference between the healthy and unhealthy people. This
motivates the idea of splitting  the whole time series into subseries.
Suppose we split the time series into $K$ subseries for each subject. 
Correspondingly we also split each mode function $F_i$ into 
$K$ subcomponents, which are denoted by
$F_{i}^k,\ k=1,\ldots, K.$ For each subcomponent  $F_{i}^k$, 
we compute the 30 statistics as above: 10 for $F_i^k$ itself, 10 for the local maxima $U_i^k,$
and 10 for the local minima $L_i^k.$ 
For each $i$ and each statistics, we have $K$ values from the $K$ subcomponents. 
We compute the mean, the first quartile (the 25th quantile),
the third quartile (the 75th quantile) of these $K$ values to obtain 3 features. 
This gives 90 features. So for each model function $F_i$ we get 120 features in total.

For physiological signals, we believe the trend and low frequency
components are determined by the fundamental mechanism while the
individual differences should be reflected by the high frequency
components. In case that we do not have much knowledge about the
disease to be diagnosed we may assume the features may also come
from the trend. So the same 120 statistics are also computed for the
trend component.

To represent these features, we introduce the notations for the statistics and 
three subscripts to indicated how the statistics is calculated. 
The detailed descriptions are listed in Table \ref{table:notation}. 

\begin{table}[h]
\begin{center}
\begin{tabular}{l|c|l}
\hline\hline
Type & Notations or Values & Description \\
\hline\hline
\multirow{8}{*}{Statistics} & $m$ & mean of the time series \\
 &  $\sigma$  & standard deviation of the time series \\
 & $mm$ &  mean of subcomponent  means \\
 & $m\sigma$  & mean of subcomponent standard deviations \\ 
  & $qm$ &  1st quartile of subcomponents  means \\
 & $q\sigma$  & 1st quartile of subcomponent standard deviations \\
 & $Qm$ &  3rd quartile of subcomponents  means \\
 & $Q\sigma$   & 3rd quartile of subcomponent standard deviations \\
\hline\hline
\multirow{2}{*}{Subscript 1} & 
$i=1, 2, \ldots, m$ & Statistics computed from $F_i$. \\
 & $i=R$ & Statistics computed from $R$. \\
\hline\hline
\multirow{3}{*}{Subscript 2} & 
 $j=0$ or omitted & Statistics for the whole series or subseries.\\
&  $j=+1$ or $+2$ & Statistics for the terms greater than $m+j\sigma$.\\
&  $j=-1$ or $-2$  & Statistics for the terms  less than $m-|j|\sigma$. \\
\hline\hline
\multirow{3}{*}{Subscript 3} & 
 $0$ or omitted &  Statistics computed from $F_i$ or $R$.\\
& $L$ &  Statistics computed from local minima.\\
 & $U$ &  Statistics computed from  local maxima.\\
\hline\hline
\end{tabular}
\caption{\label{table:notation} The notations for the features.}
\end{center}
\end{table}

\subsection{Feature subset selection}

After the above two steps we get a large number of  features for the data.
Usually only a small part of them are related to the diagnosis and
the physiological mechanism of the disease. The task of the third
step is to find the relevant ones. This will be realized by
eliminating the irrelevant ones step by step.

Firstly, if a statistic is almost constant, then it is useless in
the diagnosis and should be eliminated. For example, the means of
the mode functions $m_i$ are all approximately zero and should be
eliminated.

Next we use the SVM-RFE method \citep{SVMRFE} to rank the features.
In this method, given a set of training samples, we first train
 linear SVM to get a linear classifier and then rank the features according
to the weights. Because of large feature size and small training
samples, the classifier might not be as good. Also, the high
correlation between the features may result the relevant features to
have small weights. These reasons could lead the rank to be
inaccurate. In order to refine the rank we eliminate the least
important feature and repeat the process to re-rank the remained
features. Running this process iteratively we finally get the
refined rank of the features.

With this rank of features we can conclude which statistics are
useful for the diagnosis and characterize the essence of the
underlying physiological mechanism. Good classifiers can then be
built to make accurate diagnosis.

\section{Experiments and Results}
\label{sec:experiment}

In this section we apply our new method described above to
the heart beat interval times series and report our results and
conclusions.

\subsection{The data set}

The data set includes the heart beat interval time series of 72
healthy people and 43 CHF patients. For each subject the heart beat
interval is measured for 24 hours under various activities. In our
experiment we will assume the activity period is not known. 
The CHF of $43$ patients are classified into 4 degrees where the
degree I is a slight CHF and the degree IV is a severe CHF. Most CHF
patients are of the degree III.

\subsection{A primary study}\label{sec:primary}

\begin{figure}
\begin{center}
\includegraphics[width = .7\textwidth]{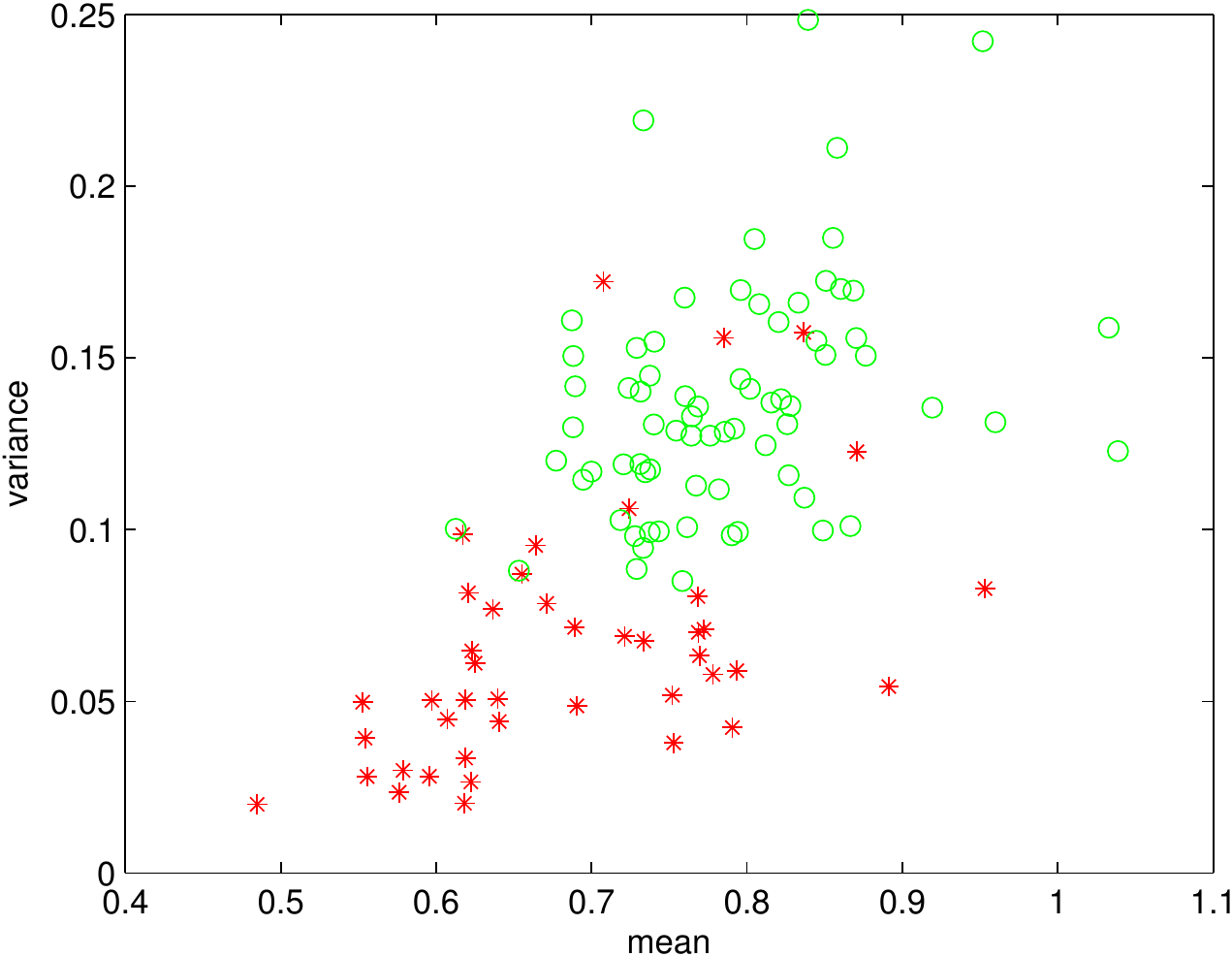} 
\end{center}
\caption{The mean and variance (in second) of heart beat interval times series, `o'
for healthy subjects and `*' for CHF patients. \label{fig:meanvar} }
\end{figure}

Before using our new method, we study the classification ability of
two simple statistics: mean and variance. In Figure
\ref{fig:meanvar} we plot the mean and variance of the heart beat
intervals for the healthy people and CHF patients. We see that the
healthy people and the CHF patients can be roughly separated. The
average heart beat interval of healthy people is larger and so is
the variance. It shows the heart of healthy people beats slower and
more irregularly. This observation is consistent with the previous studies.

At the same time, we notice that several CHF patients  falling into the
cluster of healthy people show to be severe CHF patients. So we conjecture that
the mean and variance might not reflect the essence of the
underlying mechanism, although they have good separability.

\subsection{Experiment: feature extraction}

For each time series, we use the iterative convolution filter to
realize the signal decomposition. In this step we need to specify
the window size of the mask. It turns out it should be chosen
between 50 and 100 to be stable. In our experiment it is chosen to
be $50.$

We then calculate the statistics proposed in Section \ref{sec:feature}. Here we
need to specify the parameter $K$, the number of subseries.  If a
statistic really captures the essence of the data set, it should be
stable and independent of the choice of $K$ once it is chosen within
a reasonable interval. Our experiments show that $K=50$ is a good
choice. Most heart beat signals were recorded for a little bit more
than 24 hours. Thus when $K=50$, each subseries is around 30 minutes
of record.

Previous studies have shown that healthy heart beats irregularly. In
statistics, irregularity could be measured by statistics of
``outliers" that are not due to noise. This motivates us to consider
the statistics of upward and downward fluctuations. At
the same time, from the study in Section \ref{sec:primary} we find
that a healthy heart beats slower than an unhealthy heart in
average. These two intuitions enlighten us to conjecture that those
larger heart beat intervals (i.e. slower heart beats) in the time
series characterize the difference between the healthy people and
CHF patients. To confirm this, we do a correlation analysis.

\begin{figure}[htbp]
\begin{center}
\includegraphics[width=\textwidth]{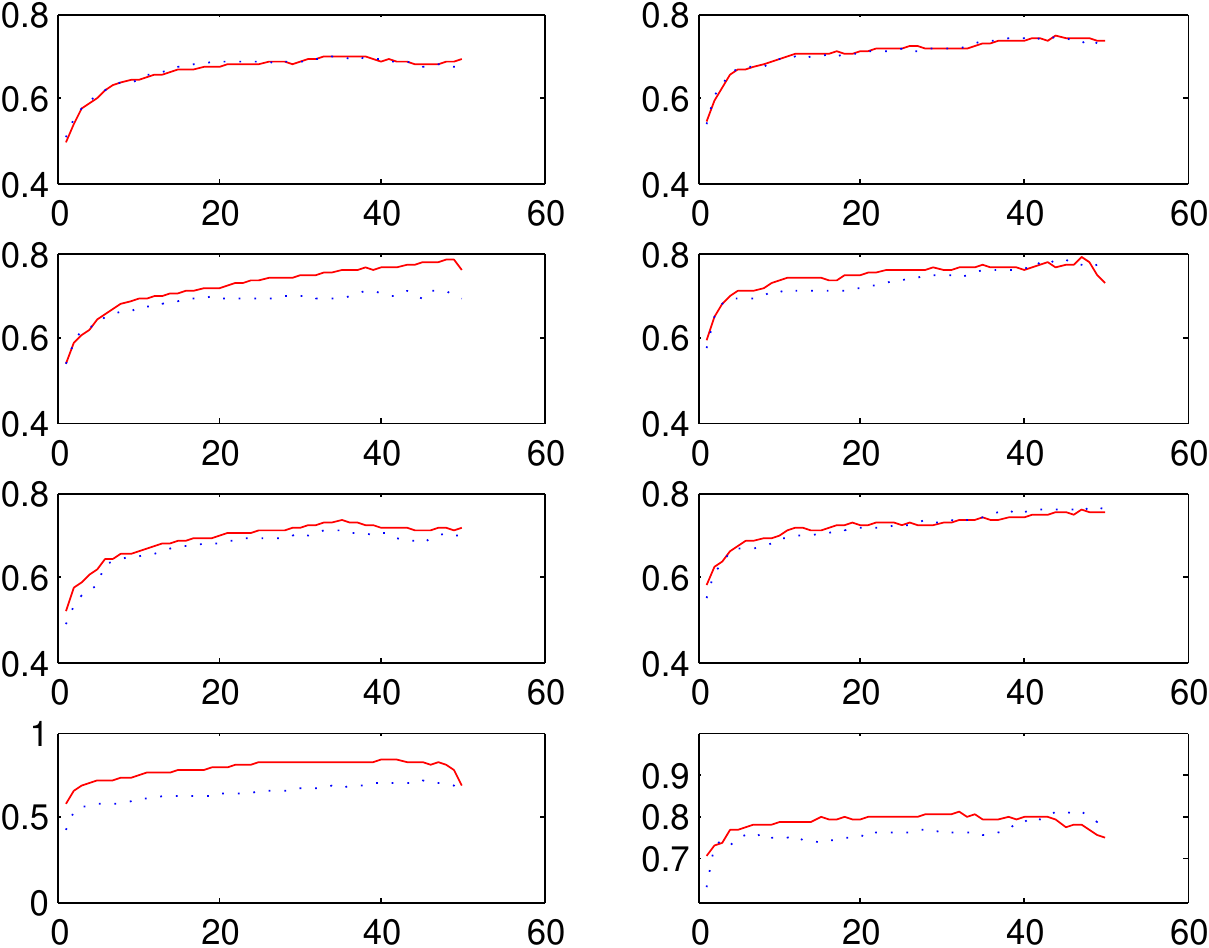}
\end{center}
\caption{The correlations of various statistics to the CHF disease.
The first column is for the first mode function  $F_1$  and the second column is for
the second mode function $F_2$. 
The first row is for order statistics of $m_{i,1}^k$ (red solid line) versus the order statistics of $m_{i,-1}^k$ (blue dotted line).
The second row is for order statistics of $\sigma_{i,1}^k$ (red solid line) versus the order statistics of $\sigma_{i,-1}^k$ (blue dotted line).
The third row is for order statistics of $m_{i,2}^k$ (red solid line) versus the order statistics of $m_{i,-2}^k$ (blue dotted line).
The last row is for order statistics of $\sigma_{i,2}^k$ (red solid line) versus the order statistics of $\sigma_{i,-2}^k$ (blue dotted line).
\label{fig:updown}}
\end{figure}

For each of the first two mode functions and each $j=1,2,-1,-2$, 
we calculate and sort the means $m_{ij}^k$ and standard deviations $\sigma_{ij}^k$
for the $K=50$ subcomponents.  For each order statistics we
compute its correlation to the CHF disease. The result is plotted in Figure \ref{fig:updown}. 
From the comparison we see that, in average,
correlations of the statistics associated to upward fluctuations
are larger than those associated to downward fluctuations.
This observation tells that we may disregard the statistics for the downward fluctuations.

\subsection{Feature ranking and subset selection}

To rank the features, we randomly split the data set into two
subsets as the training set and the test set, respectively. In the
training set we have 50 healthy subjects and 30 CHF subjects and in
the test set there are 22 healthy and 13 CHF subjects. We use the
training set to build the SVM classifier and use the test set to
control  the accuracy. Using the SVM-RFE methods described in
Subsection 2.3 we rank the features. To guarantee the stability of
the rank we repeat this procedure 1000 times and choose the
statistics that appear most frequently in the model.

In all 1000 repeats, the classification error on the test data set
is summarized in the following table:
\begin{table}[ht]
\begin{center}
\begin{tabular}{c | c |c|c| c|c | c } \hline\hline
 number of errors & 0 & 1 & 2 & 3 & 4 & 5 \\ \hline
 number of repeats & 823 & 116 & 42 & 14 & 4 &1 \\ \hline\hline
\end{tabular}
\label{tbl01}
\caption{Number of errors and the corresponding number of repeats.}
\end{center}
\end{table}

\begin{figure}[htbp]
\begin{center}
\includegraphics[width=0.8\textwidth]{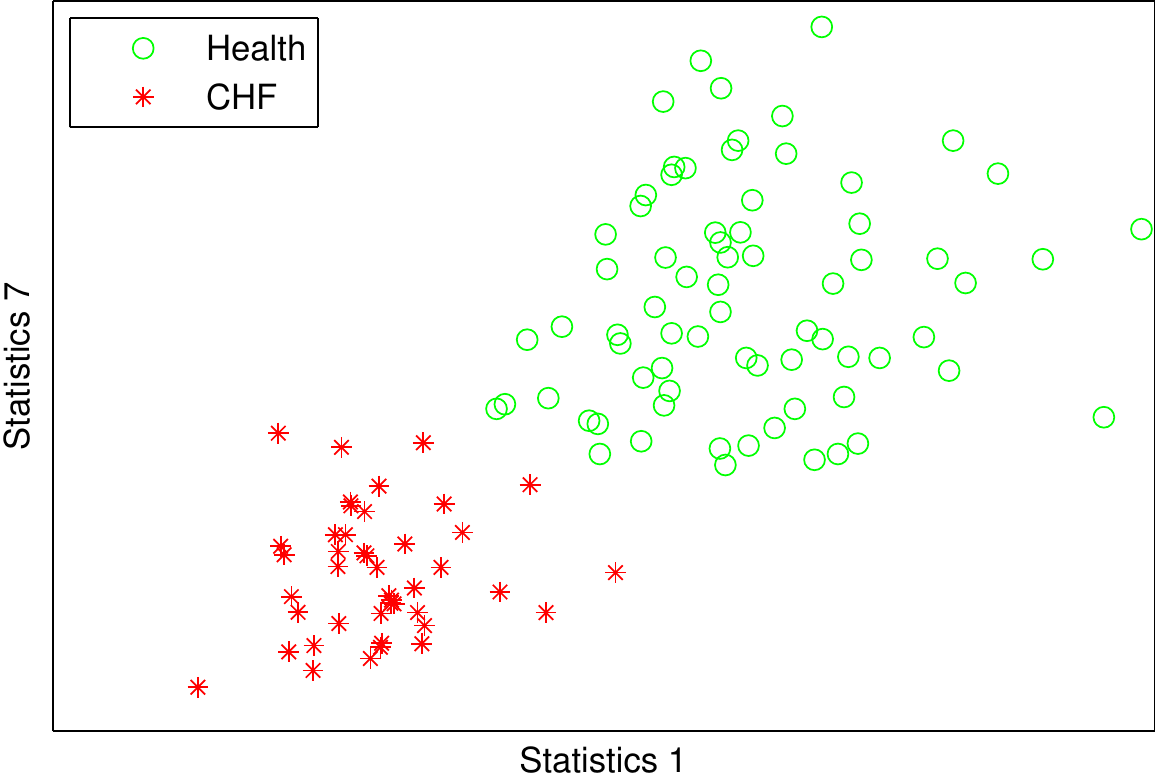}
\caption{ Distribution of CHF patients vs healthy subjects using Statistics 1 ($\sigma_{1,2,U}$) and
 Statistics 7 ($m\sigma_{2,2}$). \label{fig01}} 
\end{center}
\end{figure}

The top 10 features selected by the procedure are listed in Table \ref{table:topfeatures}.
We see 9 of them are related to the first two IMFs. 
Although the trend is in general not considered relevant,
the last feature, associated to the trend, appears. We suspect 
a  probable reason is that using only two mode functions in the signal decomposition leaves 
some relevant information in the trend. 
It is interesting to notice that these 10 statistics that appear most frequently in the model all
measure the irregularity of the local amplitude. 
Take Statistics 1
and Statistics 7 as the example. They are obtained as the following.
To get Statistics 1, for the first mode function $F_1$, find the local maxima
$U_1$ and compute its mean $m_{1,0,U}$ and the standard deviation $\sigma_{1, 0,U}$. 
Then we choose terms greater than $m_{1,0, U}+2\sigma_{1,0,U}$ and find their
standard deviation. To get Statistics 7, for the subcomponents of the
second mode function, $F_{2}^k, k=1,\ldots,K$, compute the mean $m_{2}^k$ and the
standard deviation $\sigma_{2}^k$. Then we choose terms
greater than $m_{2}^k+2\sigma_{2}^k$ of $F_{2}^k$ and find their
standard deviations $\sigma_{2,2}^k.$ 
Then we compute the mean of $K$ such standard
deviations. In Figure \ref{fig01} we show the distribution of the
healthy people and CHF patients using these two statistics.
It is easy to see that healthy people and CHF patients
are well separated.

\begin{table}[h]
\begin{center}
\begin{tabular}{l|c|c|c|c|c}
\hline\hline
Feature Rank & 1 & 2 & 3 &4 &5 \\ \hline
Statistics & $\sigma_{1, 2, U}$ & $\sigma_{1, -2,  U}$ & 
$m\sigma_{1, 2}$ & $m\sigma_{1, 2, U}$ &
$m\sigma_{1, 2, L}$  \\
\hline\hline 
Feature Rank &6& 7& 8& 9 & 10 \\ \hline 
Statistics &  $\sigma_{2,2}$ & 
$m\sigma_{2,2}$  & $m\sigma_{2, 2, U}$ &
$m\sigma_{2, -2, L}$  & $m\sigma_{R, 1, U}$ \\
\hline\hline
\end{tabular}
\caption{\label{table:topfeatures} The top 10 features.}
\end{center}
\end{table}

Observing these two statistics, we find that both of them measure
the ability of the heart beat to become extremely slower than usual.
It leads to the conjecture that the strong adaptability of extremely slower
heart beat might be the irregularity that characterizes the healthy
hearts.

\subsection{Reliability of the top features}
\label{sec:stability}

We have found that the most relevant features are statistics for the
``outliers" in the mode functions, i.e., those items larger than mean plus two times
standard deviations, or items less than mean minus two times
standard deviations. A natural question arises: ``Is this
accidental?"  This is equivalent to ask whether the outliers taken
into account are noise or informative.

In order to answer this question we further analyze these outliers.
Firstly we notice that the up and down fluctuations are not balanced
for both healthy people and CHF patients. The percentage of items
larger than mean plus two times standard deviation for healthy
people is 2.84\% and those items smaller than the mean minus two stand
deviation is only 2.35\%. For CHF patients the percentages are
2.49\% and 2.17\%, respectively. This observation is the first
evidence that outliers are not due to noise because otherwise they
should be balanced distributed. Moreover, recall for Gaussian white noise
the percentage of one-side outliers outside the two
times standard deviation is 2.28\%. We see the outliers for CHF is
closer to it 
while those for healthy subjects are much larger.
We think that the outliers for CHF patients  involve more noise 
while the outliers for healthy subjects are probably informative.

\begin{figure}[h]
\begin{center}
\includegraphics[width=0.6\textwidth]{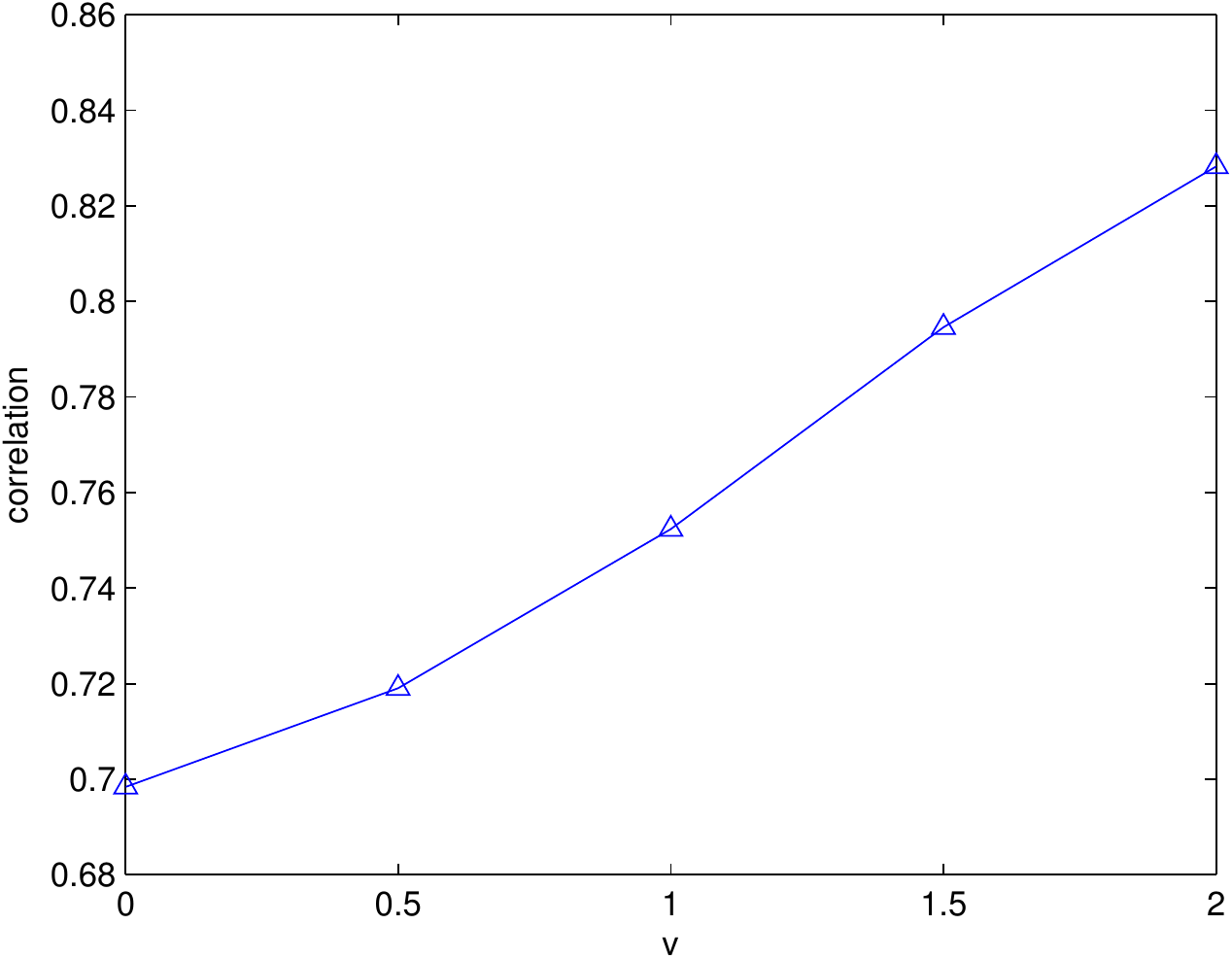}
\end{center}
\caption{Correlations between CHF disease and 
the mean of the 50 standard deviations of those terms
greater than the mean plus $v$ times standard deviation 
in the 50 subcomponents.  The value of $v$ varies from 0 to 2.\label{fig:stable}}
\end{figure}

To further confirm our conclusion, we do the following test: for $F_1$, we
calculate the statistics for the terms greater than the mean plus $v$
times standard deviation with the variable $v$ changing from 0 to 2
and investigate their correlation to the CHF disease. Here we
consider mean of the 50 standard deviations of such terms
in the 50 subcomponents. Note Statistics 3 in Table \ref{table:topfeatures} 
corresponds to $v=2.$
The correlation is plotted in Figure
\ref{fig:stable}. From this analysis, we see the correlation
increases with $v$. Such a trend appears also in other statistics.
This clear trend implies that the relevancy between these statistics
and the CHF disease is not accidental. Instead, we should consider
the outliers informative and their properties characterize the
essence difference between healthy people and CHF patients.

\section{Conclusions and discussions}

In this paper we developed a new approach for the analysis of the
physiological times series. The motivation comes from that the
physiological times series usually contains both deterministic and
stochastic parts and they can be represented by the low and high
frequency components of the times series. Our new method uses an
iterative filter to realize the decomposition of the times series
into high and low frequency components and study their statistics.
SVM-RFE is then used to select highly relevant features.

Our method is applied to analyze the heart beat interval time series
for CHF disease. The top features are found to measure the ability
of hearts to beat extremely slowly. Healthy hearts show strong ability
which we conjecture is due to the strong resilience to the
environment and human activities.

\section*{Acknowledgement}

Y. Wang was partially supported by NSF DMS-1043032 and AFOSR FA9550-12-1-0455.

\bibliographystyle{abbrvnat}
\bibliography{chfbib}

\begin{thebibliography}{23}
\providecommand{\natexlab}[1]{#1}
\providecommand{\url}[1]{\texttt{#1}}
\expandafter\ifx\csname urlstyle\endcsname\relax
  \providecommand{\doi}[1]{doi: #1}\else
  \providecommand{\doi}{doi: \begingroup \urlstyle{rm}\Url}\fi

\bibitem[Casolo et~al.(1989)Casolo, Balli, Taddei, Amuhasi, and
  Gori]{casolo1989decreased}
G.~Casolo, E.~Balli, T.~Taddei, J.~Amuhasi, and C.~Gori.
\newblock Decreased spontaneous heart rate variability in congestive heart
  failure.
\newblock \emph{The American journal of cardiology}, 64\penalty0 (18):\penalty0
  1162--1167, 1989.

\bibitem[Chen et~al.(2006)Chen, Huang, Riemenschneider, and Xu]{Chen}
Q.~Chen, N.~Huang, S.~Riemenschneider, and Y.~Xu.
\newblock A {B}-spline approach for empirical mode decompositions.
\newblock \emph{Adv. Comput. Math.}, 24\penalty0 (1-4):\penalty0 171--195,
  2006.

\bibitem[Costa et~al.(2005)Costa, Goldberger, and Peng]{med1}
M.~Costa, A.~L. Goldberger, and C.-K. Peng.
\newblock Multiscalce entropy analysis of biological signals.
\newblock \emph{Physical Review, E}, 71:\penalty0 021906, 2005.

\bibitem[Echeverria et~al.(2001)Echeverria, Crowe, Woolfson, and
  Hayes-Gill]{Echeverria}
J.~Echeverria, J.~Crowe, M.~Woolfson, and B.~Hayes-Gill.
\newblock Application of empirical mode decomposition to heart rate variability
  analysis.
\newblock \emph{Medical and Biological Engineering and Computing}, 39:\penalty0
  471--479, 2001.

\bibitem[Guyon et~al.(2002)Guyon, Weston, Barnhill, and Vapnik]{SVMRFE}
I.~Guyon, J.~Weston, S.~Barnhill, and V.~Vapnik.
\newblock Gene selection for cancer classification using support vector
  machines.
\newblock \emph{Machine Learning}, 46:\penalty0 389--422, 2002.

\bibitem[Huang et~al.(2009)Huang, Yang, and Wang]{Wang2}
C.~Huang, L.~Yang, and Y.~Wang.
\newblock Convergence of a convolution-filtering-based algorithm for empirical
  mode decomposition.
\newblock \emph{Advances in Adaptive Data Analysis}, 1\penalty0 (04):\penalty0
  561--571, 2009.

\bibitem[Huang et~al.(1998)Huang, Shen, Long, Wu, Shih, Zheng, Yen, Tung, and
  Liu]{Huang}
N.~E. Huang, Z.~Shen, S.~R. Long, M.~C. Wu, H.~H. Shih, Q.~Zheng, N.~Yen,
  C.~Tung, and H.~H. Liu.
\newblock The empirical mode decomposition and the {H}ilbert spectrum for
  nonlinear and non-stationary time series analysis.
\newblock \emph{Proceedings of the Royal Society of London, A}, 454:\penalty0
  903--995, 1998.

\bibitem[Huang et~al.(1999)Huang, Shen, and Long]{Huang2}
N.~E. Huang, Z.~Shen, and S.~R. Long.
\newblock A new view of nonlinear water waves: the {H}ilbert spectrum.
\newblock In \emph{Annual review of fluid mechanics, {V}ol.\ 31}, pages
  417--457. Annual Reviews, Palo Alto, CA, 1999.

\bibitem[Hughes et~al.(2012)Hughes, Mao, Rockmore, Wang, and
  Wu]{hughes2012empirical}
J.~M. Hughes, D.~Mao, D.~N. Rockmore, Y.~Wang, and Q.~Wu.
\newblock Empirical mode decomposition analysis for visual stylometry.
\newblock \emph{Pattern Analysis and Machine Intelligence, IEEE Transactions
  on}, 34\penalty0 (11):\penalty0 2147--2157, 2012.

\bibitem[Ivanov et~al.(1999)Ivanov, Amaral, Goldberger, Havlin, Rosenblum,
  Struzik, and Stanley]{ivanov1999multifractality}
P.~C. Ivanov, L.~A.~N. Amaral, A.~L. Goldberger, S.~Havlin, M.~G. Rosenblum,
  Z.~R. Struzik, and H.~E. Stanley.
\newblock Multifractality in human heartbeat dynamics.
\newblock \emph{Nature}, 399\penalty0 (6735):\penalty0 461--465, 1999.

\bibitem[Jiang et~al.(2011)Jiang, Peng, and Xu]{jiang2011hierarchical}
Y.~Jiang, C.-K. Peng, and Y.~Xu.
\newblock Hierarchical entropy analysis for biological signals.
\newblock \emph{Journal of Computational and Applied Mathematics}, 236\penalty0
  (5):\penalty0 728--742, 2011.

\bibitem[Kantz and Schreiber(1997)]{Kantz2}
H.~Kantz and T.~Schreiber.
\newblock \emph{Nonlinear time series analysis}, volume~7 of \emph{Cambridge
  Nonlinear Science Series}.
\newblock Cambridge University Press, Cambridge, 1997.

\bibitem[Kantz et~al.(1998)Kantz, Kurths, and Mayer-Kress]{Kantz}
H.~Kantz, J.~Kurths, and G.~Mayer-Kress.
\newblock \emph{Nonlinear techniques in physiological time series analysis}.
\newblock Springer series in synergetics. Springer, Heidelberg, 1998.

\bibitem[Lin et~al.(2009)Lin, Wang, and Zhou]{Lin}
L.~Lin, Y.~Wang, and H.~Zhou.
\newblock Iterative filtering as an alternative algorithm for empirical mode
  decomposition.
\newblock \emph{Advances in Adaptive Data Analysis}, 1\penalty0 (04):\penalty0
  543--560, 2009.

\bibitem[Liu et~al.(2006)Liu, Riemenschneider, and Xu]{Liu}
B.~Liu, S.~Riemenschneider, and Y.~Xu.
\newblock Gearbox fault diagnosis using empirical mode decomposition and
  {H}ilbert spectrum.
\newblock \emph{Mechanical Systems and Signal Processing}, 20\penalty0
  (3):\penalty0 718--734, 2006.

\bibitem[Peng et~al.(1995)Peng, Havlin, Stanley, and
  Goldberger]{peng1995quantification}
C.-K. Peng, S.~Havlin, H.~E. Stanley, and A.~L. Goldberger.
\newblock Quantification of scaling exponents and crossover phenomena in
  nonstationary heartbeat time series.
\newblock \emph{Chaos: An Interdisciplinary Journal of Nonlinear Science},
  5\penalty0 (1):\penalty0 82--87, 1995.

\bibitem[Pines and Salvino(2002)]{Pines}
D.~Pines and L.~Salvino.
\newblock Health monitoring of one dimensional structures using empirical mode
  decomposition and the {H}ilbert-{H}uang transform.
\newblock In \emph{Proceedings of SPIE}, volume 4701, pages 127--143, 2002.

\bibitem[Poon and Merrill(1997)]{poon1997decrease}
C.-S. Poon and C.~K. Merrill.
\newblock Decrease of cardiac chaos in congestive heart failure.
\newblock \emph{Nature}, 389\penalty0 (6650):\penalty0 492--495, 1997.

\bibitem[Schreiber(1999)]{Schreiber}
T.~Schreiber.
\newblock Interdisciplinary application of nonlinear time series methods.
\newblock \emph{Phys. Rep.}, 308\penalty0 (2), 1999.

\bibitem[Tong(1990)]{Tong}
H.~Tong.
\newblock \emph{Nonlinear Time Series Analysis}.
\newblock Oxford University Press, Oxford, 1990.

\bibitem[Yang et~al.(2006{\natexlab{a}})Yang, Yang, and Qi]{Yang2}
Z.~Yang, L.~Yang, and D.~Qi.
\newblock Detection of spindles in sleep {EEG}s using a novel algorithm based
  on the {H}ilbert-{H}uang transform.
\newblock In T.~Qian, M.~I. Vai, and Y.~Xu, editors, \emph{Wavelet Analysis and
  Applications}, Applied and Numerical Harmonic Analysis, pages 543--559.
  Birkhauser, 2006{\natexlab{a}}.

\bibitem[Yang et~al.(2006{\natexlab{b}})Yang, Yang, Qi, and Suen]{Yang}
Z.~Yang, L.~Yang, D.~Qi, and C.~Suen.
\newblock An {EMD}-based recognition method for {C}hinese fonts and styles.
\newblock \emph{Pattern Recognition Letter}, 27:\penalty0 1692--1701,
  2006{\natexlab{b}}.

\bibitem[Zheng et~al.(2008)Zheng, Xie, and Yang]{Zheng}
T.~Zheng, L.~Xie, and L.~Yang.
\newblock Integrated extraction on handwritten numeral strings in form document
  based on hybrid binarization.
\newblock \emph{Pattern Recognition and Artificial Intelligence}, 21\penalty0
  (3):\penalty0 369--375, 2008.

\end{thebibliography}

\end{document}